  \documentclass{gretsi}

   \usepackage[english,french]{babel}   % "babel.sty" + "french.sty"
  \usepackage{times}			% ajout times le 30 mai 2003
  
  \usepackage{epsfig}
\usepackage{graphicx}

\title{Evaluating object detector ensembles for improving the robustness of \\artifact detection in endoscopic video streams}

\author{\coord{Pedro Esteban}{Chavarrias-Solano}{1},
        \coord{Carlos Axel}{Garcia-Vega}{1},
        \coord{Francisco Javier}{Lopez-Tiro}{1},
        \coord{Gilberto}{Ochoa-Ruiz}{1},
        \coord{Thomas}{Bazin}{2},
        \coord{Dominique}{Lamarque}{2},
        \coord{Christian}{Daul}{3}}

\address{\affil{1}{Escuela de Ingenieria y Ciencias, Tecnologico de Monterrey \\
         Av. Eugenio Garza Sada 2501 Sur, Tecnológico, 64849 Monterrey, N.L., Mexico}
         \affil{2}{ H\^opital Ambroise Paré  \\ 9 Avenue Charles de Gaulle, 92100 Boulogne-Billancourt, France}
         \affil{3}{ CRAN (UMR 7039), Université de Lorraine and CNRS,  \\ 2 avenue de la For\^et de Haye, 54518 Vandœuvre-l\`es-Nancy cedex, France}}

\email{ A00344305@tec.mx, A01754346@tec.mx, A01799045@tec.mx, gilberto.ochoa@tec.mx \\ thomasbazin@icloud.com,
dominique.lamarque@aphp.fr, christian.daul@univ-lorraine.fr}

\frenchabstract{Dans cette contribution nous utilisons une méthode ensembliste d’apprentissage profond pour combiner la prédiction de deux détecteurs individuels à un étage (c'est-à-dire YOLOv4 et Yolact) dans le but de détecter les artefacts vus dans des images endoscopiques. Cette « stratégie ensembliste » a permis d'améliorer la robustesse des modèles individuels sans nuire à leurs capacités de calcul en temps réel. L'efficacité de notre approche a été démontrée en entraînant et testant les deux modèles individuels et diverses configurations ensemblistes sur le jeu de données « Endoscopic Artifact Detection Challenge ». Des expériences poussées montrent la supériorité, en termes de précision moyenne, de l’approche ensembliste par rapport aux modèles individuels et aux travaux de l'état de l'art.}

\englishabstract{In this contribution we use an ensemble deep-learning method for combining the prediction of two individual one-stage detectors (i.e., YOLOv4 and Yolact) with the aim to detect artefacts in endoscopic images. This ensemble strategy enabled us to improve the robustness of the individual models without harming their real-time computation capabilities. We demonstrated the effectiveness of our approach by training and testing the two individual models and various ensemble configurations on the “Endoscopic Artifact Detection Challenge” dataset. Extensive experiments show the superiority, in terms of mean average precision, of the ensemble approach over the individual models and previous works in the state of the art. }

\begin{document}
\maketitle

\section{Introduction}
\label{introduction}

% Endoscopy progress in recent years, promising applications of endoscopy, the role of CV in endoscopy, applications
Endoscopy is a technique that has been widely  used over two centuries  by physicians to screen the interior of otherwise inaccessible sites in the human body \cite{Fu2021}. Nowadays, it is the primary diagnostic and therapeutic tool for managing gastrointestinal (GI) malignancies \cite{Tang2020} and the primary instrument in minimally invasive surgery (MIS) procedures \cite{Fu2021}. As the endoscope provides a high quality video signal, it has fostered the development of a great deal of tasks related to image analysis and computer vision \cite{Munzer2016}. Some of the most promising applications in this area are related to the detection of diseases or pre-cancerous lesions such as polyps, ulcer, bleeding, Celiac and Chron's disease.  Recently, the use of computer vision (CV) approaches in endoscopy has caught the attention of the artificial intelligence (AI) and medical research communities due to the advent of deep learning approaches. Typically used of computer vision in endoscopy fall into two main categories: Computer-aided detection (CADe) and computer-aided diagnosis (CADx) systems \cite{He2019}.

% The challenges that remain, the impact of artifacts, public datasets regarding this issue
Even though promising applications have been demonstrated, there are several challenges that must be addressed before AI can be successfully deployed in endoscopic interventions in real clinical settings: for instance, images can be corrupted by various types of artefacts, making object detection and instance segmentation methods less robust and unable to generalize , among other pressing issues \cite{Ali2020}. In order to tackle these problems, various public datasets have been released to develop tools capable of handling these complex settings and various competitions are organized every year in major conferences. 

For instance, the EndoCV is a crowd-sourced challenge that aims  to address these issues by developing reliable  and robust CADe/x endoscopy systems \cite{Ali2020}.
% Two stage detectors to address this challenge, their advantages and limitations
In recent editions of this challenge, the teams that obtained the two highest detection scores implemented an ensemble of one two-stage and one single-stage detectors \cite{Ali2020}. The main advantage of employing two-stage detectors is the overall improvement in terms of robustness. However, these methods have an important limitation in the clinical context, which is related to the very high inference time, as they cannot be used for CADe or CADx applications.

% State of the art
More recently, several proposals in the challenge  have been proposed to overcome these limitations. In \cite{polat2020}, an ensemble of RetinaNet, Cascade and Faster R-CNN was proposed with a class-agnostic NMS stage after each model. An ensemble of Faster R-CNN RetinaNet  and Faster R-CNN was implemented in \cite{arnavchavan2020}. On the other hand, a YOLACT implementation  with Non-\-Maxi\-mum Suppression (NMS) was proposed by \cite{xiaohong2020}.  

% Area of opportunity: Robust and executed in real time object detectors, ensemble method
Even though multiple proposals have started to tackle the robustness issues discussed above, we consider that there is still a lack of robust yet real-time capable object detector that would enable various applications for the CV community in endoscopy. A way of dealing with this problem of robustness is to adopt a model ensambling strategy, but the models need to be preferably lightweight (to avoid becoming too power hungry and memory intensive) and run in real-time \cite{juan-carlos}.

% Contribution: Ensemble technique using YOLO and YOLACT
To investigate how this can be achieved, in this paper we present a comparison between two single-stage object detectors, YOLOv4 and YOLACT, and we propose an ensemble using both of them. The main contribution consists on developing an ensemble mechanism using only single-stage detectors aiming to improve the robustness of the detection task while maintaining a low memory footprint and inference time.

% Paper organization
This paper is organized as follows: In Section II, a brief description of EndoCV challenge, YOLOv4 and YOLACT architectures is given, followed by the ensemble method that was implemented for this dataset. Section III covers the results that were obtained during this comparison and section IV concludes the article with some future avenues of research.

\section{Data and methods}
\label{data_methods}
This study compares the use of two single-stage detectors, YOLOv4 and YOLACT against an ensemble of both of them, which we dub CEM. These models were trained and tested with the Endoscopy Artifact Detection challenge 2020 \cite{Ali2020}.

\subsection{EAD Dataset}
\label{EAD_Dataset}
The EndoCV challenge consists of Endoscopy Disease Detection (EDD) and Endoscopy Artifact Detection (EAD) tasks. The EAD sub-challenge contains diverse endoscopy video frames that were collected from seven institutions. The dataset  covers eight different artefact classes that were identified by clinical experts as specularity, saturation, artefact, blur, contrast, bubbles, instrument and blood. The dataset is composed by 2,532 images, classified into two types of data: single and sequence frames \cite{Ali2020} for testing different models.

\begin{figure*}[h]
    \centering
    \includegraphics[width=0.7\linewidth, height = 5cm]{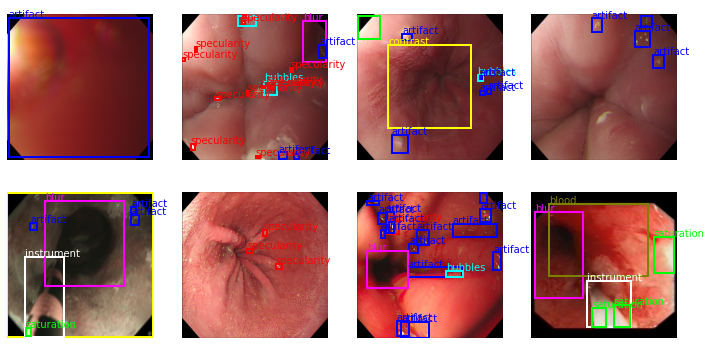}
    \caption{Samples from EAD2020 dataset images \cite{Ali2020}}
    \label{fig:EAD}
\end{figure*}

At the top of Fig. \ref{fig:EAD}, five images from the single frame category are shown. While, at the bottom of the same figure, five images belonging to the sequence frames category are shown.

\begin{figure}[h]
    \centering
    \includegraphics[width=0.8\linewidth]{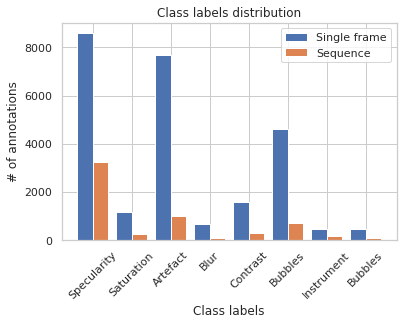}
    \caption{Class distribution from EAD2020 dataset \cite{Ali2020}}
    \label{fig:EAD_Data}
\end{figure}

For the detection task, 31,069 bounding boxes were given, being specularity, the class with more occurrences, followed by artefact and bubbles. Whereas, the class with less occurrences is blood, followed by instrument, blur, and saturation. A chart describing the data distribution is given in Fig.  \ref{fig:EAD_Data}.

\subsection{Object Detectors}
\label{Object Detectors}
The first single-stage object detector that was trained and evaluated with this dataset was YOLOv4 \cite{Bochkovskiy2020}. The architecture of this model was set to work with CSPDarknet53 as its backbone, an additional SPP module, PANet path-aggregation neck and a YOLOv3 head.

The other single-stage object detector that was selected for this study is YOLACT's lightweight instance segmentation detector \cite{dbolya2019}, whose architecture closely follows the one of RetinaNet, without the use of focal loss. We set our backbone detector to be ResNet50 with a modified FPN, applying smooth-L1 loss to train box regressors and encode box regression coordinates, as SSD.

\subsection{Ensemble}
\label{Ensemble}
An ensemble method combines the predictions done by multiple object detectors into a final output \cite{Casado-Garcia2020}. This method can be understood as a voting system in which every model in the ensemble submits its own predictions, aiming that the final decisions accuracy overcomes the accuracy of every learning method alone. 
We implemented two voting strategies: 
\\
\begin{itemize}
    \item \textbf{Consensus:} This voting strategy needs the majority of the models to detect the same object in order to consider that prediction a final output. Since we are structuring our ensemble method with just two single-stage object detectors, this approach works exactly as the unanimous voting strategy, in which both YOLOv4 and YOLACT models need to agree with the same prediction.
    \item \textbf{Affirmative:} Unlike the consensus and unanimous strategies, this approach requires just a single model, either YOLOv4 or YOLACT, to do a prediction in order to take it into account for the final output. In other words, all predictions done by each detector will be considered for the final result.
\end{itemize}

\subsection{Training}
\label{Training}
The training procedure of both object detectors, YOLOv4 and YOLACT, was performed on an NVIDIA DGX-1 system with eight Tesla V100 GPUs. Before the training procedure was done, input images were resized to 416 x 416 pixels and augmented using different data augmentation techniques: flips, blur and hue, gamma and equalized histogram and a combination of all of them.

The YOLOv4 model was trained using a learning rate of 0.001, a batch size of 64, subdivision equals to 16, momentum value of 0.949 and weight decay of 5$e^{-4}$ on a single GPU. The traditional YOLACT model was also trained using a single GPU with a learning rate of 0.001, batch size of 16 images, weight decay of 5$e^{-4}$, a momentum value of 0.9.
The ensemble method was implemented by using the generated object detector models from the training stage.

\subsection{Metrics}
The performance of the methods was evaluated using an standard evaluation metric: mean average precision (mAP). This metric measures the ability of a model to accurately capture all instances of the ground truth annotations. 

This metric was evaluated at three different intersection over union values (IoU): 0.25, 0.50, and 0.75. The IoU value measures the overlap of two different bounding boxes and is used to determine whether or not a prediction is correct with respect to the ground truth.

\section{Results}
\label{results}
This section compares qualitatively and quantitatively the obtained results after completing an evaluation of the selected models: YOLOv4, YOLACT and the ensemble method. 

\subsection{Quantitative Results}
\begin{figure}[h]
    \centering
    \includegraphics[width=0.85\linewidth]{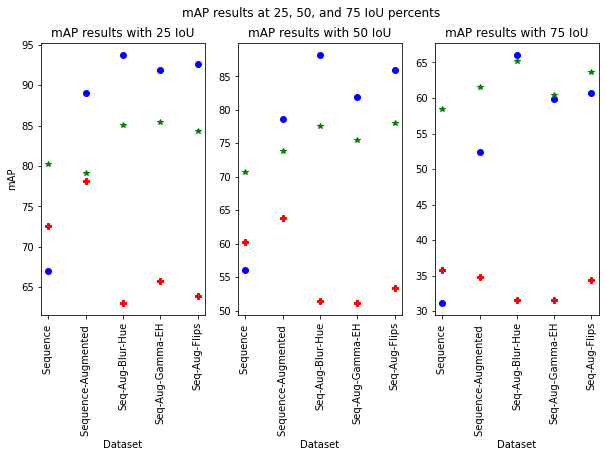}
    \caption{Mean average precision values at 25, 50, and 75 IoU values. Red crosses: YOLOv4 mAP values, Blue Circles: YOLACT mAP values and Green stars: CEM}
    \label{fig:maps}
\end{figure}

The mAP metric was used to evaluate all three methods at three different IoU: 25, 50, and 75 percent. The graphs in Fig. \ref{fig:maps} show the mAP values obtained by each model with different data augmentation strategies: original data only, geometric data augmentation, distortion (blur and hue), photo-metric (gamma and equalized histogram) and flips. The red crosses indicate the YOLOv4 mAP values, while the blue circles denote the YOLACT mAP values and finally, the green stars highlight the mAP values of the consensus ensemble method (CEM).

From the figure, we can observe that the ensemble model outperforms the other methods for at 75 IoU, while preserving a low inference time given by the two one-stage detectors.

One of the main objectives of the challenge is to address generalization issues, therefore, the test data that was released to the participants was collected from different sources not presented in the training data. In this study, the public training data was randomly split to construct the train and test datasets since official test data is not publicly available. Even though the models evaluated during the challenge and our models were not evaluated over the same data, table \ref{table:benchmark} presents the results obtained by the three best performing teams, the given baseline and the methods described in this study, i.e. YOLOv4, YOLACT and CEM, with the data augmentation that achieves the best mAP for the CEM method. From the table we can observe that all our methods outperform previous competitors challenge, as well as the baselines by a large margin. YOLACT has the highest mAP for the IoU of 25 and 50 (91.88 and 81.95 respectively), but the CEM ensembles obtained the best overall mAP for IoU of 75 (60.47).

\begin{table}[]
\caption{Benchmark between three best performing teams, given baselines \cite{Ali2020} and the methods used in this study }
\label{table:benchmark}
\begin{tabular}{llll}
\hline
\multicolumn{1}{l|}{\textbf{Team names}}   & \multicolumn{1}{l|}{\textbf{mAP 25}} & \multicolumn{1}{l|}{\textbf{mAP 50}} & \textbf{mAP 75} \\ \hline
\multicolumn{1}{l|}{StarStarG}             & \multicolumn{1}{l|}{46.965}          & \multicolumn{1}{l|}{30.202}          & 5.432           \\ \hline
\multicolumn{1}{l|}{higersky}              & \multicolumn{1}{l|}{47.716}          & \multicolumn{1}{l|}{29.841}          & 4.473           \\ \hline
\multicolumn{1}{l|}{qzheng5}               & \multicolumn{1}{l|}{48.21}           & \multicolumn{1}{l|}{25.717}          & 3.997           \\ \hline
\multicolumn{4}{l}{\textit{\textbf{baselines}}}                                                                                            \\ \hline
\multicolumn{1}{l|}{YOLOv3}                & \multicolumn{1}{l|}{32.199}          & \multicolumn{1}{l|}{18.473}          & 1.137           \\ \hline
\multicolumn{1}{l|}{RetinaNet (ResNet101)} & \multicolumn{1}{l|}{17.646}          & \multicolumn{1}{l|}{6.447}           & 0.767           \\ \hline
\multicolumn{4}{l}{\textit{\textbf{Ours}}}                                                                                                 \\ \hline
\multicolumn{1}{l|}{YOLACT}                & \multicolumn{1}{l|}{\textbf{91.88}}           & \multicolumn{1}{l|}{\textbf{81.95}}           & 59.8           \\ \hline
\multicolumn{1}{l|}{YOLOv4}                & \multicolumn{1}{l|}{65.83}            & \multicolumn{1}{l|}{51.22}           & 31.55          \\ \hline
\multicolumn{1}{l|}{CEM}                   & \multicolumn{1}{l|}{85.44}           & \multicolumn{1}{l|}{75.5}           & \textbf{60.47}            \\ \hline
\end{tabular}
\end{table}

\subsection{Qualitative Results}

\begin{figure}[h!]
    \centering
    \includegraphics[width=0.85\linewidth]{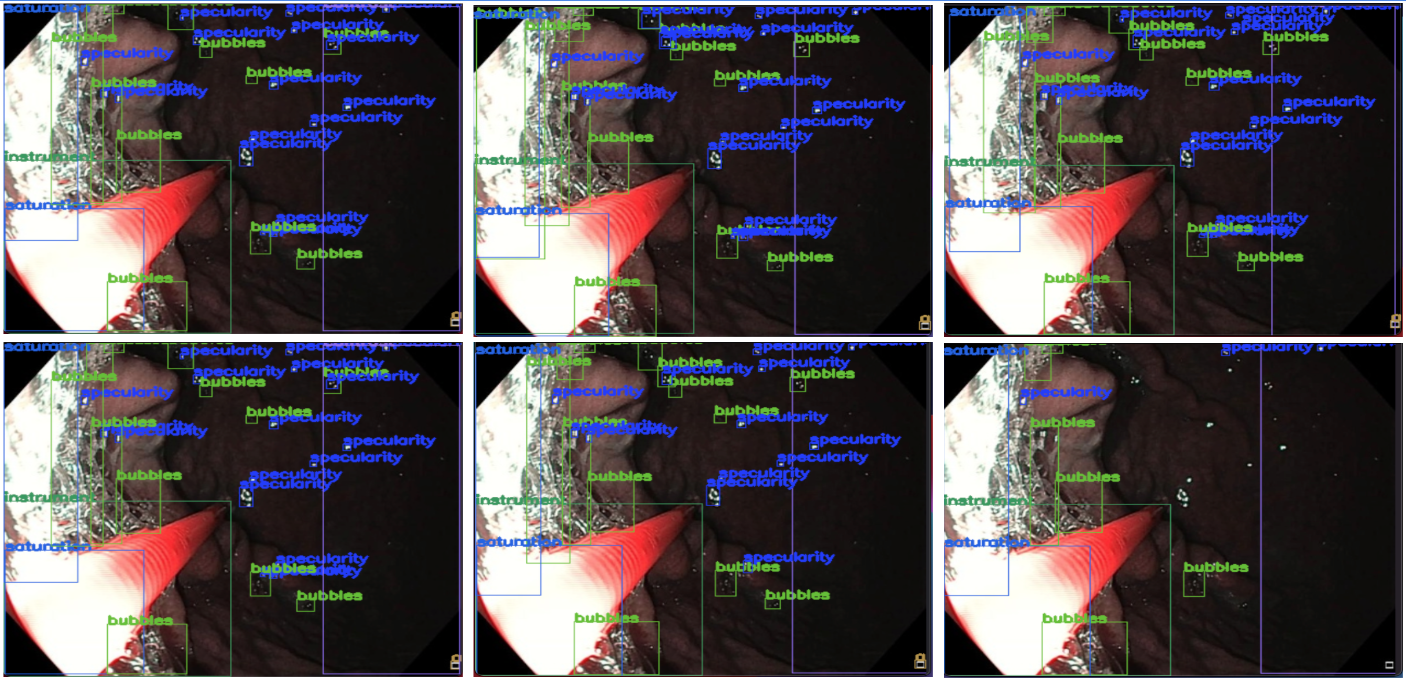}
    \caption{Visual comparison between ground truth and inference results of YOLOv4, YOLACT, CEM at 50 and 75\% IoU}
    \label{fig:Results}
\end{figure}

Figure \ref{fig:Results} contains a set of predictions that were done by the trained models over a test image. Also, the ground truth is shown in order to do a visual comparison between the expected results and the actual predictions. 
The 1st column contains the expected ground truth bounding boxes. The 2nd column shows the predictions done by YOLOv4 and the CEM with a 50\% IoU. Finally, the 3rd column displays the predictions done by YOLACT and the CEM with a 75\%  IoU.

It can be seen how as the IoU threshold is increased, the amount of final detections decrease. In fact, this situation is more clear in small objects (i.e., specularities not detected in bottom right image). Thus, more experiments and architectural improvements are needed to deal with such small objects.
\section{Conclusion and future work}
\label{discussion_future_work}
A proposed CEM was generated from two single-stage detectors, YOLOv4 and YOLACT, trained models with different data augmentations, aiming to improve the detection performance while trying to maintain inference time as low as possible. All three models were evaluated in terms of mAP at three different IoU thresholds: 25, 50, and 75. Based on the obtained results, CEM's object detector have a better performance than YOLOv4 and YOLACT at higher IoU. 

Another important aspect was that even though the ensemble method increases inference time, it outperformed YOLOv4 in all cases, which means that an ensemble of two single-stage object detectors can increase the mAP value at a little inference time cost with respect to an ensemble of two-stage detectors.

\bibliography{references}

\end{document}